\documentclass[letterpaper]{article}
\usepackage{aaai26}
\usepackage{times}
\usepackage{helvet}
\usepackage{courier}
\usepackage[hyphens]{url}
\usepackage{graphicx}
\usepackage{natbib}
\usepackage{caption}
\usepackage{xcolor}
\usepackage{amsmath}
\usepackage{amssymb}
\usepackage{booktabs}
\usepackage{algorithm}
\usepackage{algpseudocode}
\title{Dynamic Parameterization Is Not Dynamic Inference}
\author{
Zongfei Li,
Yuan-yih Shang,
Guozhong Luo
}
\affiliations{}
\nocopyright
\begin{document}
\maketitle

\begin{abstract}
Input-dependent controller coefficients are often treated as evidence of dynamic inference or computational savings. This interpretation conflates three properties: coefficient variation, dependence of a frozen model on how coefficients are assigned to inputs, and conditional execution. We focus on the second property and formulate a general principle of frozen-controller auditing. We provide one concrete implementation, \textbf{F}rozen-\textbf{C}ontroller \textbf{A}uditing (\textbf{FCA}), which caches the complete coefficient tensor along an unperturbed trajectory, disables the controller, and replays the frozen model with cross-input reassignment, token shuffling, and static profiles estimated from an independent calibration set. Because the coefficients are cached before any intervention, performance changes under replay measure assignment dependence without feedback from recomputing the controller on perturbed hidden states. Across seven independently trained 76M FeatureGate Transformers and three 504M models, static layerwise profiles retain 98.70\% and 99.43\% of the Correct-to-GlobalMean performance gap, respectively. Layer identity explains 87\% to 96\% of the coefficient variance. FeatureGate nevertheless executes every Transformer block, and its measured inference is 30.8\% slower than Dense. On the public MUDDPythia-1.4B checkpoint, cross-input reassignment and token shuffling increase NLL by 1.9067 and 2.9637, respectively. These penalties show that the model depends strongly on content-conditioned cross-layer assignment. MUDDPythia also executes every Transformer block. The results show that dynamic parameterization alone does not establish dynamic inference and that functional dynamics do not establish computational savings. Claims about dynamic models should separately report coefficient variation, functional dependence of the frozen model, and actual execution.
\end{abstract}
\section{Introduction}

Adaptive computation seeks to allocate model capacity according to the requirements of each input or token. Variable-step models and adaptive-depth networks change the number of refinement steps applied to a representation \citep{graves2016adaptive,dehghani2019universal,banino2021pondernet,hou2020dynabert}. Early-exit and layer-skipping systems terminate computation or bypass selected blocks for particular inputs \citep{teerapittayanon2016branchynet,wu2018blockdrop,elbayad2020depth,schuster2022confident}. Sparse expert and token-routing methods activate only selected experts or tokens under a computation budget \citep{shazeer2017moe,fedus2022switch,rao2021dynamicvit,raposo2024mixture}. Although these approaches are often described collectively as dynamic inference, they modify different aspects of inference. Hard routing changes which layers, tokens, or experts are executed and can therefore reduce FLOPs or latency. Soft residual gating evaluates every block and only rescales the corresponding updates. Dynamic cross-layer mixing can also preserve the full computation graph while changing how representations from different depths contribute to later computation. All of these mechanisms may produce input-dependent coefficients, but coefficient variation alone does not determine whether the computed function or the executed computation is genuinely dynamic.

Existing evaluations often blur this distinction. Gate heatmaps, coefficient variance, and cross-input differences establish that the outputs of a controller are nonconstant, but they do not establish that the frozen model requires the original correspondence between coefficients and inputs. A controller may vary substantially across inputs and tokens even when an input-independent layerwise profile reproduces almost the same inference function. Conversely, a dynamic mixing mechanism may depend critically on aligning each coefficient tensor with the content and token positions from which it was generated. Neither outcome implies computational savings when every block remains active. Therefore, claims about dynamic models require three separate questions: whether the coefficients vary, whether the frozen inference function depends on their input-specific assignment, and whether the mechanism changes the computation that is actually executed.
\begin{figure*}[t]
\centering
\includegraphics[width=\textwidth]{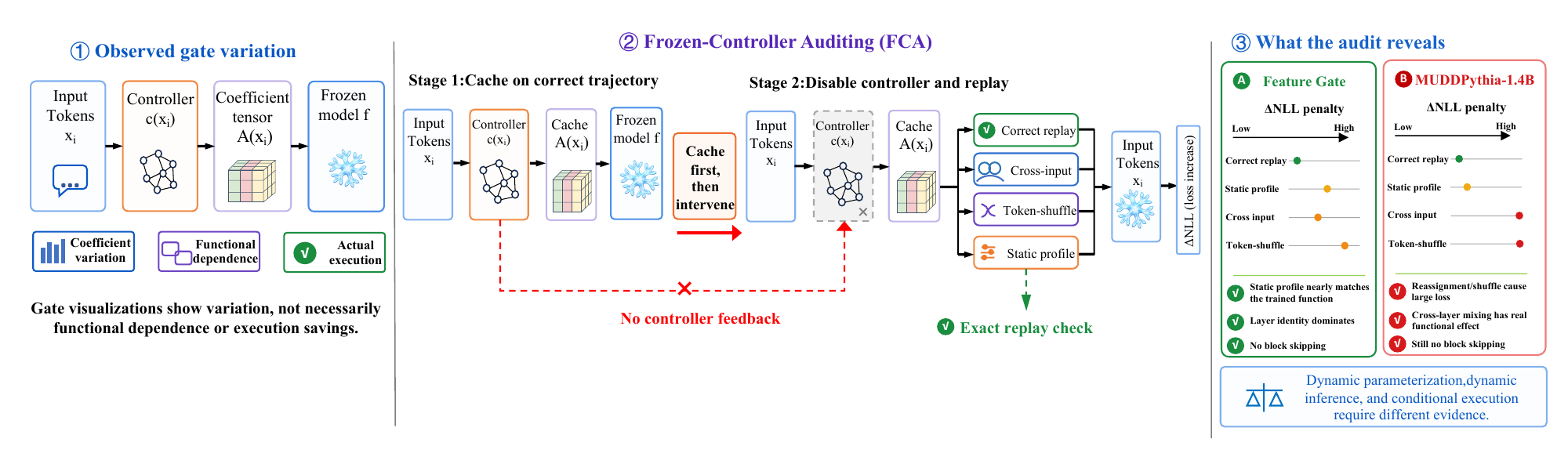}
\caption{\textbf{Overview of Frozen-Controller Auditing (FCA). } FCA distinguishes coefficient variation, functional dependence, and actual execution. It first caches controller coefficients along the unperturbed trajectory, then disables the controller and replays the frozen model with reassigned, shuffled, or static coefficients. The audit finds that a static depth profile closely reproduces the inference behavior of FeatureGate, whereas MUDDPythia depends strongly on the alignment of cross-layer coefficients with their original inputs and token positions. Both models still execute every Transformer block. Dynamic coefficients alone therefore do not establish conditional execution or computational savings.}
\label{fig:protocol}
\end{figure*}
The key insight of this work is that functional dependence can be tested directly through interventions on a frozen model. The evaluation first records the complete coefficient tensor along the unperturbed computation trajectory. It then disables the controller and replays the same model with the cached coefficients either preserved or systematically transformed. Because the coefficients are collected before any intervention and are not recomputed from perturbed hidden states, the resulting performance change isolates dependence on coefficient assignment from feedback through the controller. This idea defines a general frozen-controller auditing principle. We instantiate the principle as Frozen-Controller Audit, or FCA, using cross-input reassignment, token shuffling, static profiles estimated from an independent calibration set, and separate measurements of execution. FCA is one natural implementation of the broader principle rather than its only possible form; other coefficient-preserving interventions can be designed for different controller structures and architectural claims.

We apply FCA to two deliberately contrasting model families. FeatureGate is a soft residual gating mechanism for decoder-only Transformers \citep{vaswani2017attention}. It produces a coefficient for each layer and token but does not skip any block. Across seven independently trained 76M models and three independently trained 504M models, the outputs of FeatureGate vary across inputs and tokens, yet a static layerwise depth profile retains 98.70\% and 99.43\% of the Correct-to-GlobalMean performance gap, respectively. Layer identity explains most of the coefficient variance, and interventions that disrupt input or token assignment incur only small penalties. In contrast, the public MUDDPythia-1.4B checkpoint, which uses position-specific dynamic cross-layer mixing for the query, key, value, and residual streams \citep{xiao2025muddformer}, is highly sensitive to cross-input reassignment, token shuffling, and static replacement. The same audit therefore distinguishes a controller whose dynamic variation is largely replaceable from a controller whose dynamic assignment has a substantial functional role. Nevertheless, both model families execute every Transformer block, and the measured inference of FeatureGate is slower than that of the Dense baseline.

Our contributions are as follows:
\begin{enumerate}
    \item  We formulate a three-way evaluation principle that separates coefficient variation, functional dependence of the frozen model, and actual execution, thereby clarifying the evidence required for claims about dynamic parameterization, dynamic inference, and conditional computation.
    \item We introduce FCA as a reproducible implementation of frozen-controller intervention, combining two-stage caching and replay, coefficient-preserving reassignment, independently calibrated static profiles, factorial interaction analysis, and explicit execution measurements.
    \item We provide a multi-scale and multi-model empirical study showing that FeatureGate primarily learns a stable depth profile, whereas MUDDPythia depends strongly on content-conditioned cross-layer assignment; we further show that neither form of dynamic parameterization establishes block skipping or computational savings.
\end{enumerate}


\section{Related Work}

\subsection{Dynamic Networks and Input-Conditioned Parameterization}

Dynamic networks adjust parameters or computation paths across samples, spatial locations, or sequence positions \citep{han2022dynamic}. One branch of this literature directly generates or combines input-dependent weights. Dynamic Filter Networks generate convolutional filters from the input, HyperNetworks use one network to produce the parameters of another, and CondConv and Dynamic Convolution combine several kernels with input-dependent weights \citep{jia2016dynamic,ha2017hypernetworks,yang2019condconv,chen2020dynamicconv}. Input-conditioned parameters can be useful even when the computation graph remains dense and fixed. Dynamic parameterization therefore does not establish dynamic execution, and coefficient variation alone does not show that the trained function uses that variation.

\subsection{Adaptive Depth and Early Exit}

Adaptive Computation Time and the Universal Transformer vary the number of recurrent computation steps, while PonderNet learns a stopping distribution \citep{graves2016adaptive,dehghani2019universal,banino2021pondernet}. In vision models, BranchyNet, MSDNet, and Shallow-Deep Networks use intermediate classifiers for early exit. SkipNet and BlockDrop instead learn input-dependent block-skipping policies \citep{teerapittayanon2016branchynet,huang2018msdnet,kaya2019shallowdeep,wang2018skipnet,wu2018blockdrop}.

Transformer models adopt similar ideas. Depth-Adaptive Transformer, DeeBERT, FastBERT, PABEE, LeeBERT, BERxiT, and CALM reduce the number of executed layers through confidence criteria or learned exit rules \citep{elbayad2020depth,xin2020deebert,liu2020fastbert,zhou2020pabee,zhu2021leebert,xin2021berxit,schuster2022confident}. DynaBERT exposes subnetworks with adjustable width and depth, LayerDrop trains networks that permit depth pruning, and Mixture-of-Depths selects which tokens enter a block under a fixed capacity budget \citep{hou2020dynabert,fan2020reducing,raposo2024mixture}. Claims in this family concern execution, so accuracy alone is insufficient. The evidence must also include executed depth, FLOPs, latency, or throughput.

\subsection{Sparse Expert Routing}

Sparse MoE models expand parameter capacity while activating only a small subset of experts for each example \citep{shazeer2017moe}. GShard combines sparse routing with automatic sharding, Switch Transformer simplifies training with top-1 routing, and BASE Layers use balanced assignment to control expert load \citep{lepikhin2021gshard,fedus2022switch,lewis2021base}. V-MoE, GLaM, and Expert Choice extend sparse routing to vision, large language models, and expert-driven token assignment, respectively \citep{riquelme2021vmoe,du2022glam,zhou2022expertchoice}. These models distinguish total parameters from activated parameters, but routing distributions remain descriptive. Establishing dependence on an input-to-expert assignment requires an intervention that changes the assignment while holding expert weights and routing values fixed.

\subsection{Token-Level Dynamic Computation}

Another family of methods changes later computation at the token level. DynamicViT progressively removes unimportant visual tokens, EViT reorganizes and fuses tokens with low attention, and Dynamic Transformer chooses token resolution and exit depth according to image difficulty \citep{rao2021dynamicvit,liang2022evit,wang2021dynamictransformer}. A-ViT extends adaptive halting to spatial tokens, AdaViT jointly selects patches, attention heads, and blocks, Adaptive Token Sampling chooses an input-dependent number of tokens, and Token Merging reduces sequence length by merging tokens between layers \citep{yin2022avit,meng2022adavit,fayyaz2022ats,bolya2023tome}. Token-dependent coefficients constitute conditional computation only when they alter later execution. Stable absolute-position structure can also appear as token variation, so a suitable static baseline should preserve the joint structure of layer and position rather than only a layer mean.

\subsection{Residual Scaling and Cross-Layer Information Flow}

Stochastic Depth randomly skips residual blocks during training. Fixup, ReZero, SkipInit, LayerScale, and DeepNorm instead improve deep-network optimization through initialization or static residual scaling \citep{huang2016deep,zhang2019fixup,bachlechner2021rezero,de2020skipinit,touvron2021going,wang2022deepnet}. These methods show that layerwise coefficients can be useful without depending on the input.

Related architectures improve the flow of information across depth. DenseNet introduces dense connections; Transparent Attention, DLCL, and RealFormer reuse representations or attention scores from earlier layers; DenseFormer aggregates earlier representations through a depth-weighted average \citep{huang2017densenet,bapna2018transparent,wang2019dlcl,he2021realformer,pagliardini2024denseformer}. Hyper-Connections and MUDDFormer add dynamic connections across streams or layers \citep{zhu2025hyperconnections,xiao2025muddformer}. Our focus is not a new connection structure. We test whether a dynamic connection remains functionally necessary after training or can be replaced by a static structure.

\subsection{Intervention-Based Functional Analysis}

The presence of a component, variation in its activation, or concentrated attention does not make the component functionally necessary. Attention-head ablations show that many visible heads can be removed at inference with little effect on performance \citep{michel2019heads}. Causal mediation analysis, interchange intervention training, and causal tracing use internal interventions to identify the components and representations on which model behavior depends \citep{vig2020causal,geiger2022iit,meng2022rome}. FCA follows the same principle of testing function through intervention rather than inferring it from correlation. Dynamic controllers require two additional controls: coefficients must be collected before the trajectory is perturbed, and each intervention should preserve the coefficient multiset or stable structure whenever possible. These controls separate controller feedback from the effect of coefficient assignment.

\section{Method: Frozen-Controller Auditing}

\subsection{Problem Definition}

Let a controller map input $x$ to a coefficient tensor
\begin{equation}
A(x)=\{\alpha_{\ell,t,s,k}(x)\},
\end{equation}
where $\ell$ indexes the target layer, $t$ the token position, $s$ the computation stream, and $k$ an optional source-layer index. Scalar residual gating is the special case with one stream and no source-layer mixture. MUDDFormer uses the full tensor over target layers, tokens, streams, and source layers.

We distinguish three properties. Coefficient variation asks whether $A(x)$ changes with the input or position. Functional dependence asks whether the output of a fixed model depends on the correspondence between coefficient values and their original input coordinates. Actual execution asks whether the model skips layers, tokens, or experts and whether this change reduces FLOPs, latency, or throughput cost. FCA primarily tests functional dependence. System measurements address actual execution separately.

\subsection{Two-Stage Caching and Replay}

The first stage records coefficients along the unmodified model trajectory:
\begin{equation}
A_i=A(x_i;\ \text{Correct trajectory}).
\end{equation}
The second stage disables the controller and replays the model with coefficients transformed by $T$:
\begin{equation}
\mathcal{L}_T=\mathcal{L}\!\left(x_i;\ \text{Replay}(T(A_i))\right).
\end{equation}
We define the intervention penalty as
\begin{equation}
\Delta\mathrm{NLL}(T)=\mathcal{L}_T-\mathcal{L}_{\text{Correct}}.
\end{equation}
Because $A_i$ is fixed before the intervention, $\Delta\mathrm{NLL}$ excludes feedback in which perturbed hidden states change the controller output again. Every audit begins with an exact replay of the original coefficients. The intervention results are valid only if this replay reproduces the original logits and NLL.

\begin{algorithm}[t]
\caption{\textbf{F}rozen-\textbf{C}ontroller \textbf{A}udit (\textbf{FCA})}
\label{alg:fca}
\begin{algorithmic}[1]
\Require Frozen model $f$, controller $c$, evaluation set $\mathcal{D}$, independent calibration set $\mathcal{C}$, intervention family $\mathcal{T}$
\State Run the unmodified model on $\mathcal{C}$ and estimate static profiles that preserve layer, position, stream, and source-layer structure
\For{each evaluation example $x_i\in\mathcal{D}$}
    \State Run $(f,c)$ along the Correct trajectory and cache the full coefficient tensor $A_i$
    \State Disable $c$, replay $A_i$ without modification, and verify exact agreement in logits and NLL
    \For{each transformation $T\in\mathcal{T}$}
        \State Construct a cross-input reassignment, token shuffle, or static replacement $T(A_i)$
        \State Replay $T(A_i)$ in the same frozen model and record per-example NLL and execution statistics
    \EndFor
\EndFor
\State Compute paired $\Delta\mathrm{NLL}$ values, factor effects, the interaction effect, static retention, and execution measures
\end{algorithmic}
\end{algorithm}

\subsection{Assignment Interventions and Static Profiles}

The Correct condition returns the cached coefficients of each input to their original coordinates. Cross-input applies a global derangement across examples. It preserves coordinate structure and the complete coefficient multiset but changes the input from which each tensor originates. Token-shuffle independently permutes token positions within the relevant layers and computation streams. It preserves coefficient values and support while breaking token alignment. Cross+shuffle applies both transformations, with paired sampling of donor mappings and position permutations.

Layer$\times$PositionMean averages coefficients over examples in the independent calibration set while preserving target layer, computation stream, absolute position, and source-layer structure. It removes content dependence but retains stable position patterns. StaticProfile also averages over token positions. For FeatureGate, this condition is the layerwise LayerMean. For MUDDPythia, it is StaticMixMean over target layer, computation stream, and source layer.

StructureRemoved provides a model-specific reference that removes the structure of interest. FeatureGate uses a single GlobalMean. MUDDPythia uses SourceUniformMean, which removes source-layer preference while preserving the signed coefficient sum for each target layer and stream. These references have different mathematical meanings and support only within-model interpretation. They do not provide a causal ranking across the two model families.

The matching constraints above apply to controller coefficients, coefficient means, or signed sums. An effective residual update also depends on hidden states and block outputs. Coefficient matching is therefore not exact matching of residual updates unless the intervention also controls those update vectors.

\subsection{Factor Effects and Static Retention}

Let $CC$ denote correct input and token assignments, $DC$ a changed input donor with token order preserved, $CS$ the correct input with shuffled tokens, and $DS$ a changed input donor with shuffled tokens. The paired factor effects are
\begin{align}
E_{\text{input}} &= \tfrac{1}{2}\left[(DC-CC)+(DS-CS)\right],\\
E_{\text{token}} &= \tfrac{1}{2}\left[(CS-CC)+(DS-DC)\right],\\
E_{\text{interaction}} &= DS-DC-CS+CC.
\end{align}
These quantities are factorial contrasts, not an additive decomposition into independent mechanisms. A negative interaction means that the two interventions damage overlapping functions or otherwise combine nonadditively.

For a static profile $S$ and the model-specific structure-removal reference $R$, we define
\begin{equation}
\mathrm{Retention}(S)=1-\frac{\Delta\mathrm{NLL}(S)}{\Delta\mathrm{NLL}(R)}.
\end{equation}
Retention must be reported with the absolute $\Delta\mathrm{NLL}$ and interpreted only within the same model and evaluation set.

\section{Experiments and Analysis}

The experiments address four questions in sequence: (a) whether an input-conditioned controller outperforms static parameter sharing when gating strength is matched; (b) whether the frozen FeatureGate model depends on dynamic coefficient assignment; (c) whether the same audit detects strong functional dependence in an external model; and (d) whether these dynamic mechanisms reduce the computation actually executed during inference. We discuss each result immediately after the experiment that addresses the corresponding question and do not place the results in a separate section.

\subsection{Models and evidence protocol}

The 76M FeatureGate model is a decoder-only Transformer with 24 layers, a hidden dimension of 512, and 8 attention heads, totaling 75.9M parameters. We train seven instances with independent random seeds and evaluate their byte-level language modeling performance on OWT and WT103 \citep{gokaslan2019openwebtext,merity2017pointer}. The residual update is
\[
h_{\ell+1,t} = h_{\ell,t} + g_{\ell,t}(x)\, F_\ell(h_\ell)_t.
\]
Dense fixes $g=1$; Uniform learns a single global coefficient; LayerStatic learns one coefficient per layer; and FeatureGate generates input-conditioned coefficients from token features. The FeatureGate controller uses attention entropy, the KL divergence between the attention distributions of adjacent layers, a normalized log-partition statistic, and the relative update norm as inputs. All soft-gated models execute every network block. The 504M extension has 40 layers, a hidden dimension of 1024, and 16 attention heads. We train three instances with independent random seeds. This experiment tests whether the mechanism changes with scale. Its training setup is not intended to establish compute optimality at approximately 500M parameters.

For the external control, we use the public \texttt{Caiyun-AI/MUDDPythia-1.4B} checkpoint at commit \texttt{3c241c8}. The model has 24 layers and 1.420B parameters and produces signed dynamic cross-layer mixing coefficients for the query, key, value, and residual streams. We estimate the static profiles of FeatureGate and MUDDPythia from 512 WT103 validation windows and evaluate them on 512 nonoverlapping WT103 test windows. Each randomized condition uses 20 paired reassignments. For FeatureGate, each training seed is an independent unit of inference. MUDDPythia has only one public checkpoint, so its intervals describe stability across evaluation windows and randomization error for that checkpoint rather than uncertainty across independently trained checkpoints.

Before any intervention, each audit verifies that replaying the original coefficients reproduces the unperturbed model outputs. The primary FeatureGate analysis uses checkpoint-level bootstrap confidence intervals and exact two-sided sign-flip tests. We apply Holm correction to the three factorial effects. We specify two criteria before running the audit. We consider the LayerMean effect practically small if the upper bound of its 95\% confidence interval is below 0.002 NLL. We consider static retention high if the lower bound of its interval exceeds 95\%.

\subsection{Dynamic Parameterization and Training Performance}

This experiment tests whether an input-conditioned controller provides better language-modeling performance than static parameter sharing. Without a stable advantage, variation in the controller output is not necessary to reach the observed accuracy.

\begin{table}[t]
\centering
\footnotesize
\setlength{\tabcolsep}{3.5pt}
\caption{Main training comparison. Lower Byte-PPL is better. OWT uses held-out validation data, and WT103 uses the complete official test split. Values are the mean and standard deviation across training seeds.}
\label{tab:train}
\begin{tabular}{@{}llrlr@{}}
\toprule
Setting & Method & $n$ & Byte-PPL & Gate mass \\
\midrule
OWT 76M & Uniform & 7 & $3.7405 \pm 0.0515$ & 0.6100 \\
OWT 76M & FeatureGate & 7 & $3.7437 \pm 0.0503$ & 0.6102 \\
\midrule
WT103 76M & LayerStatic & 7 & $3.0842 \pm 0.0061$ & 0.5971 \\
WT103 76M & FeatureGate & 7 & $3.0862 \pm 0.0064$ & 0.6006 \\
WT103 76M & Uniform & 7 & $3.0899 \pm 0.0055$ & 0.6000 \\
\midrule
WT103 504M & FeatureGate & 3 & $2.6520 \pm 0.0046$ & 0.5952 \\
WT103 504M & LayerStatic & 3 & $2.6547 \pm 0.0031$ & 0.5953 \\
WT103 504M & Uniform-Matched & 3 & $2.6560 \pm 0.0031$ & 0.5953 \\
WT103 504M & Uniform-0.60 & 3 & $2.6574 \pm 0.0031$ & 0.6000 \\
\bottomrule
\end{tabular}
\end{table}

Table~\ref{tab:train} shows that FeatureGate does not outperform the mass-matched Uniform baseline on OWT at 76M scale. The two methods have nearly identical gate mass, so the result cannot be attributed to a smaller residual budget for FeatureGate. On WT103 at 76M scale, LayerStatic outperforms Uniform for all seven seeds, whereas FeatureGate does not consistently outperform LayerStatic. WT103 therefore benefits from a nonuniform depth profile, but a single learned coefficient per layer captures this benefit. Per-token coefficient generation is not required.

At 504M scale, FeatureGate has the lowest Byte-PPL and outperforms mass-matched Uniform for all three seeds, but the average gap is only 0.004000 Byte-PPL. The advantage over LayerStatic is smaller. Dynamic parameterization may provide a slight optimization benefit during training, even when a static profile closely approximates the resulting function. The next experiment tests this distinction by intervening after the controller is frozen.

\subsection{Dynamic Assignment Dependence of FeatureGate}

This experiment tests whether the trained FeatureGate function depends on dynamic assignment. The audit fixes coefficient values and then breaks their correspondence with inputs or tokens, so the performance change measures the functional role of that correspondence.

\begin{figure*}[t]
\centering
\includegraphics[width=2.0\columnwidth]{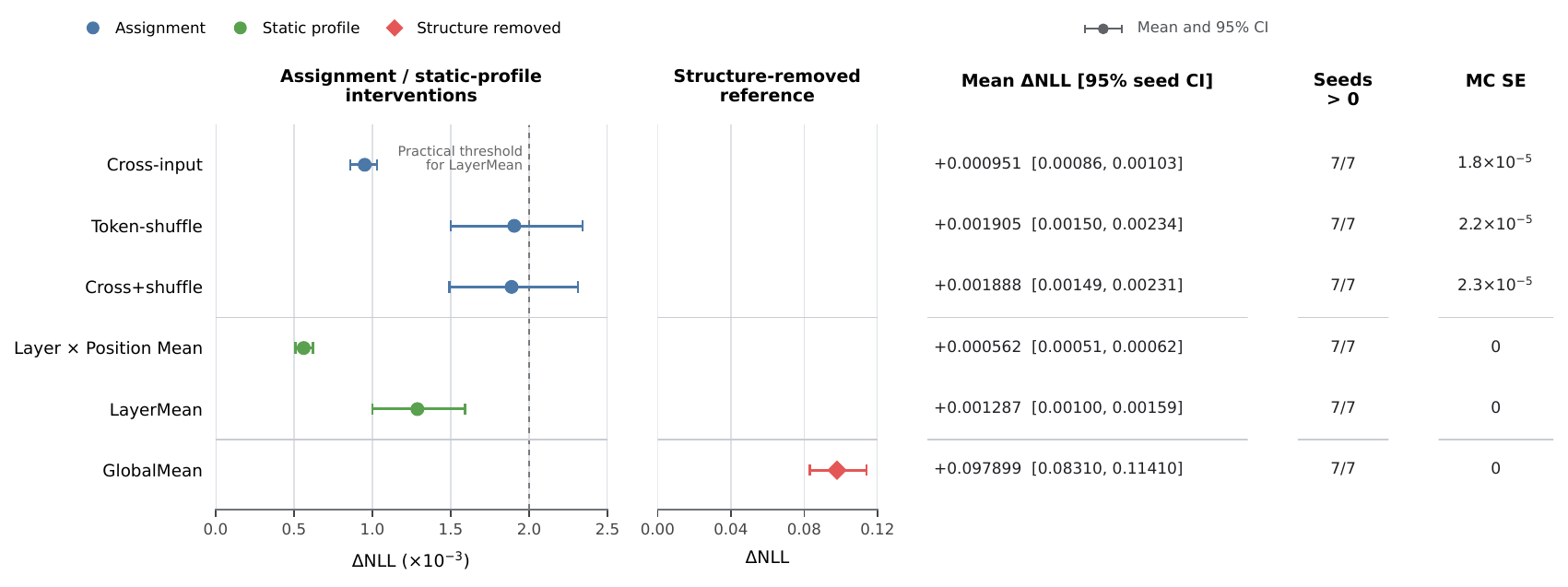}
\caption{\textbf{Frozen-Controller Audit results for 76M-parameter FeatureGate models on WT103.} Each point shows a mean intervention penalty. Horizontal error bars give $95\%$ confidence intervals across seven independently trained checkpoints. Assignment and static-profile interventions increase NLL by only about $10^{-3}$. GlobalMean removes the learned depth structure and produces a much larger penalty, so we plot it on a separate linear scale. All intervention effects are positive for every seed. The MC SE values report within-checkpoint randomization error.}
\label{fig:feature}
\end{figure*}


Every intervention increases NLL for all seven seeds, so input identity and token alignment are not completely irrelevant. The effects are nevertheless small. Cross-input increases NLL by 0.000951, and Token-shuffle increases it by 0.001905. Replacing all dynamic gates with one layerwise profile estimated from independent validation data increases NLL by only 0.001287. This value satisfies the prespecified threshold for a practically small effect and retains 98.70\% of the performance gap from Correct to GlobalMean.

The contrast between LayerMean and GlobalMean explains what the static result preserves. Both conditions remove input and token variation, but only LayerMean retains the depth profile. Preserving this profile costs about 0.0013 NLL, whereas flattening it costs nearly 0.098 NLL. The model depends strongly on where residual mass is placed across depth and only weakly on the input-dependent and token-dependent variation around that profile. Layer$\times$PositionMean reduces the penalty further to 0.000562 NLL. Some apparent token adaptivity therefore comes from a stable absolute-position pattern rather than input content.

\begin{figure}[t]
\centering
\includegraphics[width=0.82\columnwidth]{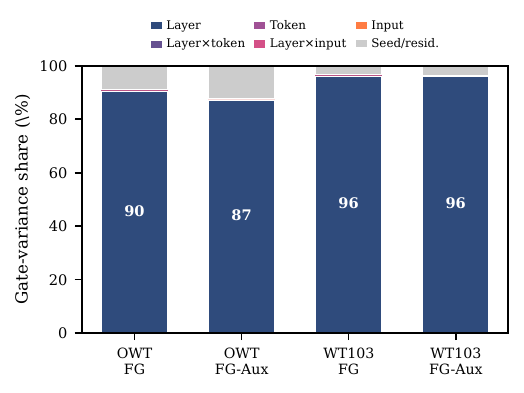}
\caption{Balanced functional variance analysis of FeatureGate coefficients. Layer identity explains 87\% to 96\% of the variance across corpora and controller variants. The input, token, and interaction terms are each below 1\%.}
\label{fig:variance}
\end{figure}

The variance decomposition provides separate descriptive evidence. Layer identity explains 87\% to 96\% of coefficient variance, while the input and token effects are small. This analysis does not replace the functional interventions, but it agrees with them: the controller primarily learns stable layerwise levels. The 504M audit follows the same pattern. Cross-input, Token-shuffle, and LayerMean increase NLL by 0.000891, 0.001574, and 0.001177, respectively, whereas the mass-matched GlobalMean increases NLL by 0.205301. The static layerwise profile retains 99.43\% of relative performance. Scaling increases the cost of destroying the fixed depth profile, not the cost of removing dynamic assignment.

\begin{figure}[t]
\centering
\includegraphics[width=0.82\columnwidth]{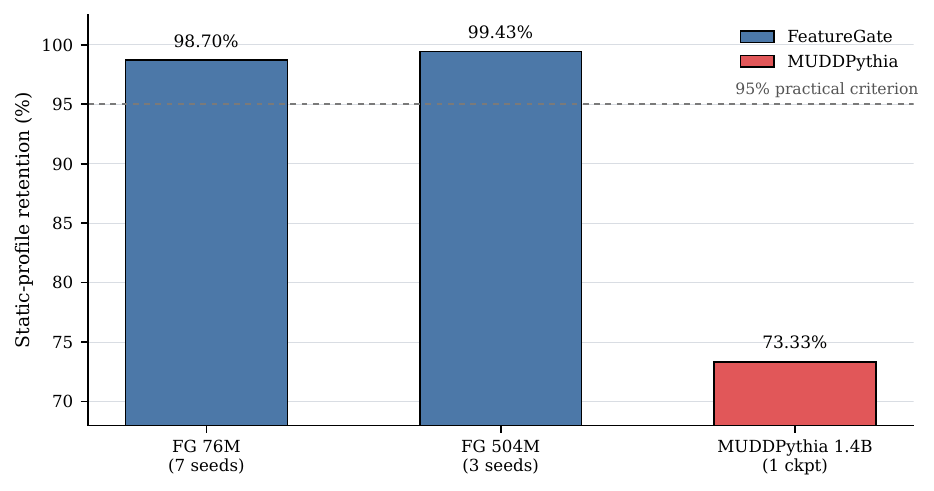}
\caption{\textbf{Static profile retention for FeatureGate and MUDDPythia.} FeatureGate LayerMean retains 98.70\% and 99.43\% of the Correct-to-GlobalMean performance gap at 76M and 504M, respectively. Horizontal bars show 95\% confidence intervals across independently trained checkpoints. For MUDDPythia, StaticMixMean retains 73.33\% of the gap relative to SourceUniformMean. This value is a point estimate from a single checkpoint. The corresponding absolute penalties are reported with each retention value. Because the two model families use different references for structure removal, retention should be interpreted within each family rather than used for direct comparison across models.} 
\label{fig:retention}
\end{figure}

The paired factor effects are $E_{\text{input}}=0.000466$, $E_{\text{token}}=0.001421$, and $E_{\text{interaction}}=-0.000968$. The negative interaction indicates that cross-input reassignment and token shuffling disrupt overlapping functions rather than two large, independent mechanisms whose costs add. The absolute effects, retention ratios, position control, and variance decomposition support a narrow conclusion: FeatureGate is dynamically parameterized, but its trained inference function is nearly static.

\subsection{Strong Functional Dependence in MUDDPythia}

The small effects for FeatureGate admit another explanation: the audit itself may be insensitive to dynamic dependence. We test this possibility by applying the same caching, replay, reassignment, and static-profile protocol to the public MUDDPythia-1.4B checkpoint.

\begin{table}[t]
\centering
\footnotesize
\setlength{\tabcolsep}{3pt}
\caption{FCA results for the public MUDDPythia-1.4B checkpoint. Intervals are descriptive paired-window bootstrap intervals. MC SE is estimated from 20 paired randomizations and does not represent checkpoint-level population uncertainty.}
\label{tab:fca-mudd}
\begin{tabular}{@{}lrrl@{}}
\toprule
Condition & NLL & $\Delta$NLL & Uncertainty \\
\midrule
Correct & 2.8661 & $0$ & N/A \\
Cross-input & 4.7728 & $+1.9067$ & MC SE 0.0021 \\
Token-shuffle & 5.8298 & $+2.9637$ & MC SE 0.0040 \\
Cross+shuffle & 5.8720 & $+3.0059$ & MC SE 0.0045 \\
Layer$\times$Pos. Mean & 3.8851 & $+1.0190$ & CI $[0.982, 1.057]$ \\
StaticMixMean & 4.8459 & $+1.9798$ & CI $[1.940, 2.022]$ \\
SourceUnif. Mean & 10.290 & $+7.4237$ & CI $[7.389, 7.458]$ \\
\bottomrule
\end{tabular}
\end{table}

The effect scale differs sharply from FeatureGate. Cross-input increases NLL by 1.9067 while preserving the full coefficient multiset, and Token-shuffle increases it by 2.9637. These penalties are roughly three orders of magnitude larger than the corresponding effects for FeatureGate. The coefficients determine which earlier-layer representations enter the query, key, value, and residual streams. Their values therefore remain useful only when aligned with the content and positions that produce them.

Static replacement also causes large losses. StaticMixMean increases NLL by 1.9798. Layer$\times$PositionMean retains absolute-position structure and reduces the penalty to 1.0190, which shows that MUDDPythia also contains stable position patterns. Those patterns do not remove the dependence on content. SourceUniformMean increases NLL by 7.4237, so source-layer preference is itself an important structure. StaticMixMean and Layer$\times$PositionMean retain 73.33\% and 86.27\% relative to SourceUniformMean. These ratios coexist with large absolute NLL penalties and cannot by themselves support the claim that most function is preserved.

The factor effects are $E_{\text{input}}=0.974430$, $E_{\text{token}}=2.031426$, and $E_{\text{interaction}}=-1.864572$. The negative interaction again indicates nonadditive overlap. More importantly, replaying the original cached coefficients exactly matches both NLL and the logits of the first batch under the normal forward pass. The large effects therefore do not arise from an inaccurate replay implementation. The same protocol distinguishes the two model families: dynamic variation in FeatureGate is largely replaceable by a static profile, whereas dynamic cross-layer assignment in MUDDPythia has a substantial functional role.

\begin{figure}[t]
\centering
\includegraphics[width=0.92\columnwidth]{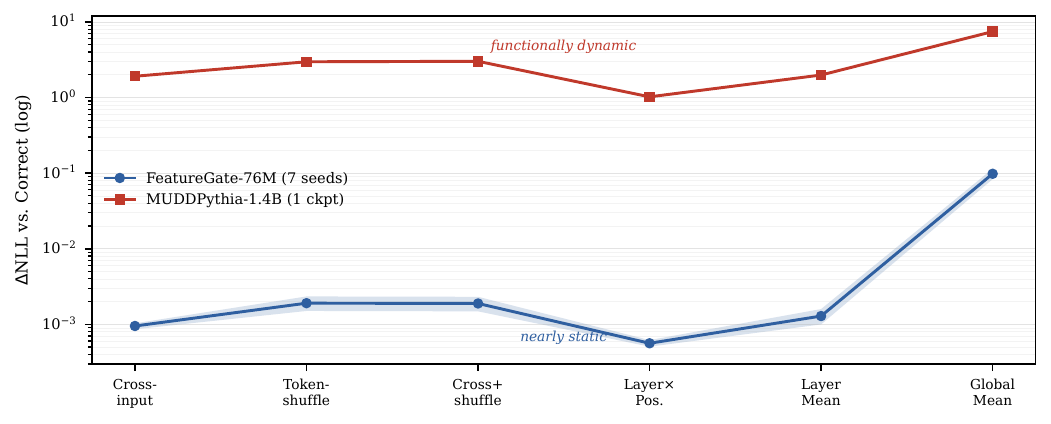}
\caption{Absolute intervention penalties relative to Correct on a logarithmic axis. Assignment and static-profile penalties remain near $10^{-3}$ NLL for FeatureGate, while the penalties for MUDDPythia are larger by one to three orders of magnitude.}
\label{fig:penalties}
\end{figure}

\subsection{Dynamic Function and Actual Execution}

Functional dynamics do not imply conditional execution, so the final experiment directly measures executed blocks, latency, and throughput. Every 76M soft-gated model executes all 24 Transformer blocks. With batch size 8 in the current dense PyTorch implementation, Dense takes 19.36 ms per batch. Uniform, LayerStatic, and FeatureGate take 23.00, 24.10, and 25.31 ms, respectively. FeatureGate is 30.8\% slower than Dense and has 23.5\% lower throughput because its controller adds computation without removing any block.

The FixedDepth15 baseline executes 15 blocks and reduces latency by 36.7\%, but it also loses accuracy. This comparison shows that actual skipping must appear in both the execution graph and system measurements. Average gate values cannot establish it. Post-training thresholding of FeatureGate produces meaningful block skipping only when PPL degrades severely, which further shows that gate mass is a residual-coefficient statistic rather than a compute measure.

MUDDPythia also executes every Transformer block. Its large FCA penalties show that dynamic cross-layer mixing has a functional role, but they do not provide evidence of block skipping. FeatureGate therefore exhibits coefficient variation but little functional dependence and no conditional execution. MUDDPythia exhibits coefficient variation and strong functional dependence, but still no conditional execution.

\section{Discussion}

The two model families show why dynamic has no single inference meaning. The controller of FeatureGate varies across inputs and tokens, and dynamic parameterization may offer a small optimization benefit at the larger scale. Yet a static depth profile estimated from independent data nearly reproduces the trained function. MUDDPythia, by contrast, depends strongly on the correspondence between coefficients, content, and position. A small static-profile penalty indicates that controller variation is largely replaceable in the frozen inference function, rather than that the controller is constant. A large penalty establishes a functional role for the dynamic connection, but it provides no evidence of computation savings. Studies of dynamic architectures should therefore report controller variation, frozen-model interventions, and actual execution statistics, with separate evidence for each claim.

\section{Limitations}

Population-level inference across seven training seeds covers only the FeatureGate family in one byte-level decoder-only Transformer. The 504M study uses three seeds and a training schedule that is not compute optimal, so it supports only the direction of the scaling result. OpenWebText has no official untouched test split, and the strongest repeated-randomization inference therefore uses WikiText-103. External validation covers one MUDDPythia checkpoint and evaluates it on WikiText-103 rather than its pretraining distribution, so the result does not generalize to all MUDDFormer training runs. The MUDDPythia experiment evaluates dynamic cross-layer mixing rather than block skipping and therefore validates only the sensitivity of FCA to functional dynamics. Static profiles match coefficient structure, means, or signed sums, but they do not exactly match effective residual updates, which also depend on hidden states and block outputs. Finally, the StructureRemoved conditions have different mathematical definitions for the two model families and do not support cross-model causal ranking.

\section{Conclusion}

Variation in controller outputs has no unique functional interpretation. For FeatureGate, a static layerwise profile nearly reproduces the trained inference function, assignment interventions cause only about $10^{-3}$ NLL loss, and soft gating does not reduce actual execution. For the public MUDDPythia-1.4B checkpoint, the same audit finds large losses under cross-input reassignment, token shuffling, and static mixing, which shows that the model materially depends on content-conditioned cross-layer connections. FCA provides a diagnostic rather than a verdict on dynamic controllers. It is a reproducible implementation of the frozen-controller auditing principle. Reliable claims about dynamic models require separate measurements of coefficient variation, functional dependence, and actual execution.

\bibliographystyle{aaai26}
\bibliography{refs}

\end{document}